\documentclass[runningheads]{llncs}
\usepackage{graphicx}
\usepackage{verbatim}
\usepackage{xspace}
\usepackage{amsmath}
\newcommand{\ra}[1]{\renewcommand{\arraystretch}{#1}}
\newcommand{\mex}{MEx\xspace}
\usepackage{multirow}
\usepackage{subfig}
\begin{document}

\title{MEx: Multi-modal Exercises Dataset for Human Activity Recognition}

\author{Anjana Wijekoon\inst{1}, Nirmalie Wiratunga\inst{1}, Kay Cooper\inst{2}}

\institute{School of Computing and Digital Media, Robert Gordon University, Aberdeen AB10 7GJ, Scotland, UK\\
           \email{a.wijekoon@rgu.ac.uk}
           \and
           School of Health Sciences, Robert Gordon University, Aberdeen AB10 7GJ, Scotland, UK\\ }

\maketitle

\begin{abstract}
MEx: Multi-modal Exercises Dataset is a multi-sensor, multi-modal dataset, implemented to benchmark Human Activity Recognition(HAR) and Multi-modal Fusion algorithms.
Collection of this dataset was inspired by the need for recognising and evaluating quality of exercise performance to support patients with Musculoskeletal Disorders(MSD).
We select 7 exercises regularly recommended for MSD patients by physiotherapists and collected data with four sensors a pressure mat, a depth camera and two accelerometers.
The dataset contains three data modalities; numerical time-series data, video data and pressure sensor data posing interesting research challenges when reasoning for HAR and Exercise Quality Assessment.
This paper presents our evaluation of the dataset on number of standard classification algorithms for the HAR task by comparing different feature representation algorithms for each sensor.
These results set a reference performance for each individual sensor that expose their strengths and weaknesses for the future tasks.
In addition we visualise pressure mat data to explore the potential of the sensor to capture exercise performance quality.
With the recent advancement in multi-modal fusion, we also believe MEx is a suitable dataset to benchmark not only HAR algorithms, but also fusion algorithms of heterogeneous data types in multiple application domains.
\end{abstract}

\section{Introduction}
Musculoskeletal Disorders (MSD) are recognised as a primary contributor to disease burden in developed countries~\cite{abajobir2017global}. 
Finding technological solutions for either the prevention or self-management of MSD have been a research area which has emerged over the last few years.
Maintaining a regular self-managed exercise routine while adhering to correct execution is an essential component when living with MSD.
An effective digital intervention should be capable of interception, recognition and quality assessment of exercises in real-time. 

Exercises can be viewed as a sub category of human activities that comprises of complex sequences of human movements.
Capturing  these  movements  with  sensors  require  much  more  detail  that  a  single accelerometer on a smart watch can provide.
They require capturing multiple limb movements and need to be captured with strategically placed multiple sensors. 
Recent literature has looked at sensor fusion for HAR with multiple sensors yet single modality~\cite{yao2017deepsense,radu2016towards,ordonez2016deep}, but not in a qualitative manner.
Evidently there are challenges that arise when reasoning with multi-modal sensors of heterogeneous data types. Sensor fusion focused on exploiting multiple sensors for improved recognition performance, non-invasive HAR and open-ended HAR for improved deployability and qualitative HAR are to name a few.

We identify the need for a heterogeneous sensory dataset for physiotherapy exercises in order to address these challenges. Importantly we collaborated with the physiotherapy researchers to identify exercises and sensors that are effective for MSD and implemented a data collection task. Outcome of this task is the MEx: Multi-modal Exercises dataset that is presented with this report. In summary, \mex consists of seven exercises recorded with four sensors, a pressure mat, a depth camera and two accelerometers. It is publicly available to download at Mendeley Data repository~\footnote{https://data.mendeley.com/datasets/p89fwbzmkd/2}. 

This paper is organised as follows. Section~\ref{sec:mex} presents the exercises, sensors and the processes followed in the data collection. Section~\ref{sec:vis} explore different data visualisation techniques with the pressure mat sensor data, followed by Section~\ref{sec:eval} and Section~\ref{sec:results} where details of experiments and results can be found for the single sensor recognition task. Finally Section~\ref{sec:conc} will present the conclusions and plans for future work. 

\section{Multi-modal Exercise Dataset (\mex)}
\label{sec:mex}
In this section we present the list of exercises, sensor specifications and the data collection methodology.

\subsection{Exercises}
Table~\ref{tbl:exercises} presents the exercises that were selected by a physiotherapist for this data collection. They are frequently used for prevention or management of Musculoskeletal pain. Each exercises is annotated with a starting position and an action; action is comprised of several guidelines and steps to accurately perform the exercise.

\begin{table}[t!]
\caption{Exercises}
\label{tbl:exercises}
\centering
\ra{1.1}
\begin{tabular}{l l}
\hline
\multicolumn{2}{l}{Knee-rolling}\\
\hline
Starting Position&Lying on back, knees together and bent, feet flat on floor\\
Action&Slowly roll knees to the right, back to the centre, then to the left,\\
&keeping upper trunk still\\
\hline
\multicolumn{2}{l}{Bridging}\\
\hline
Starting position&Lying on back with knees bent and slightly apart, feet flat on floor \\
&and arms by side\\
Action&Squeeze buttock muscles and lift hips off floor. Hold approximately\\
& 5-seconds and lower slowly.\\
\hline
\multicolumn{2}{l}{Pelvic tilt}\\
\hline
Starting Position&Lying on back with knees bent and slightly apart, feet flat on floor\\
& and arms by side\\
Action&Tighten stomach muscles and press small of back against the floor \\
&letting your bottom rise. Hold approximately 5 seconds then relax.\\
\hline
\multicolumn{2}{l}{The Clam}\\
\hline
Starting Position&Lying on right side with hips and shoulders in straight line. Bend \\
&knees so thighs are at 90 degrees angle. Rest head on top arm\\
& (stretched overhead or bent depending on comfort). Bend top arm \\
&and place hand on floor for stability. Stack hips directly on top of \\
&each other (same for shoulders)\\
Action&Keep big toes together and slowly rotate leg in hip socket so \\
&the top knee opens. Open knee as far as you can without disturbing \\
&alignment of hips. Slowly return to starting position\\
\hline
\multicolumn{2}{l}{Repeated Extension in Lying}\\
\hline
Starting Position&Lying face down, place palms on floor and elbows under shoulders \\
&(press-up position)\\
Action&Straighten elbows as far as you can and push top half of body up as \\
&far as you can. Pelvis, hips and legs must stay relaxed. Maintain \\
&position for approximately 2-seconds then slowly lower to starting \\
&position.\\
\hline
\multicolumn{2}{l}{Prone punches}\\
\hline
Starting Position&On all 4’s with hands directly beneath shoulders, knees slightly apart \\
&and straight back.\\
Action&Punch right arm in front and lower to floor. Repeat with left arm. \\
&Keep trunk as still as possible\\
\hline
\multicolumn{2}{l}{Superman}\\
\hline
Starting Position&On all 4’s with hands directly beneath shoulders, knees slightly apart \\
&and straight back.\\
Action&Extend right arm straight in front of you and left leg straight behind \\
&you, keeping trunk as still as possible. Hold approximately 5-seconds\\
& then lower and repeat with other arm and leg.\\
\hline
\end{tabular}
\end{table}

\subsection{Sensors}
Exercises are repeated controlled movements of multiple parts of the body. We explored the state of the art sensor technologies and selected three sensors to capture these movements; Obbrec Astra Depth Camera~\footnote{https://orbbec3d.com/product-astra-pro/}, Sensing Tex Pressure Mat~\footnote{http://sensingtex.com/sensing-mats/pressure-mat/} and Axivity AX3 3-Axis Logging Accelerometer~\footnote{https://axivity.com/product/ax3}. 

The aim is to explore their capabilities to capture exercises independently as well as an ensemble while considering the non-invasive use in the real world. Accordingly we select following placements for sensors; two accelerometers will be placed on the wrist and the thigh of the user; the pressure mat will be used as a exercise mat where the user will lay on to perform exercises; the depth camera will be placed above the user facing down-words recording an aerial view. In addition, top of the depth camera frame will be aligned with the top of the pressure mat and the user is asked to align their  shoulders such that the face is not recorded in the depth camera or pressure mat data. 
The four sensors, accelerometer on the thigh, accelerometer on the wrist, pressure mat and the depth camera will be referred as ACT, ACW, PM and DC in the rest of this paper. 

\begin{table}[ht]
\caption{Sensors}
\label{tbl:sensors}
\centering
\ra{1.1}
\begin{tabular}{l l}
\hline
\multicolumn{2}{l}{Depth Camera Sensor}\\
\hline
Product&Obbrec Astra Depth Camera\\
Frame rate&$\approx$15 fps\\
Frame Resolution&320$\times$240\\
\hline
\multicolumn{2}{l}{Pressure sensor}\\
\hline
Product&Sensing Tex Pressure Mat\\
Frame rate&$\approx$75 fps\\
Frame Resolution&32$\times$16\\
\hline
\multicolumn{2}{l}{Accelerometer}\\
\hline
Product&Axivity AX3 3-Axis Logging Accelerometer\\
Sample rate&100Hz\\
Accelerometer Range&$\pm$8 g\\
\hline
\end{tabular}
\end{table}

\subsection{Data Collection Methodology}
\begin{figure}[t!]
\centering
\includegraphics[width=1.0\textwidth]{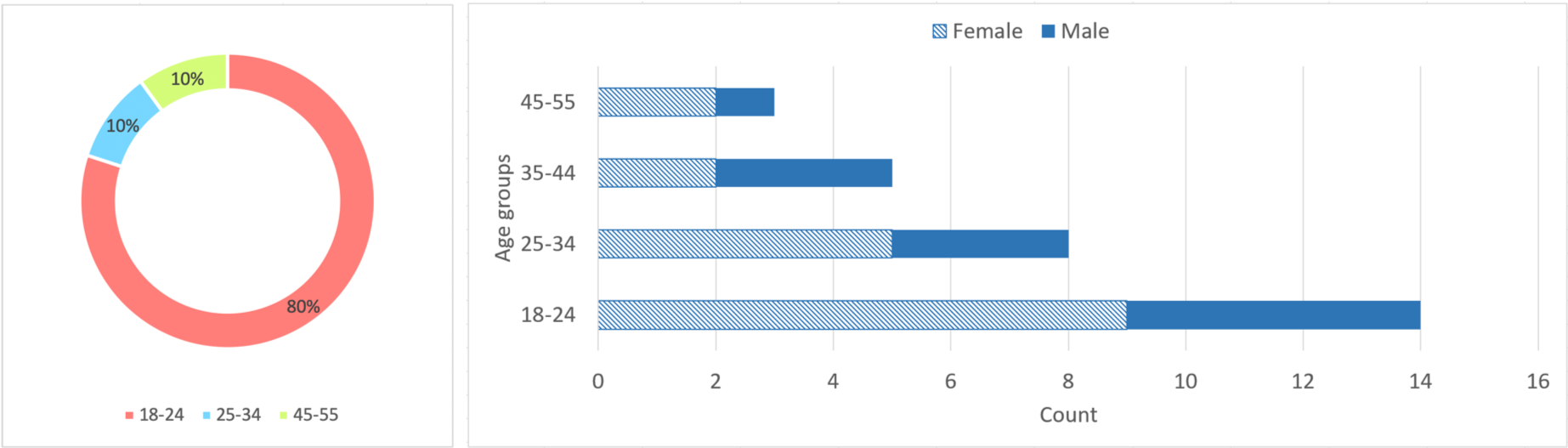}
\caption{User Demographics, Left: Percentage of physiotherapy users in each age group, Right: Gender distribution in each age group} \label{fig:users}
\end{figure}

The data collection task was performed with 30 volunteers. Figure \ref{fig:users} show the age and sex statistics of the group. 60\% of the population was female and 40\% was male. 47\% of the group were in the 18-24 age category and the rest were dispersed among ages from 24 to 54. 
The volunteers were recruited through internal adverting in the university and majority of the participants were students from the Schools of Computing or School of Health Sciences. 8 of the 30 had a good understanding of the exercises as they were either physiotherapists or physiotherapy students. 
A Physical Activity Readiness  Questionnaire (PAR-Q) was given to each user prior to participating in the study to evaluate their physical fitness and only who passed PAR-Q performed the exercises.

At the beginning of each exercise the user was given a sheet with instructions for the exercise. The document includes the starting position and the action and the researcher demonstrated each exercise to the user. Then the user performed the exercise for maximum of 60 seconds while being recorded with multiple sensors. During the recording, the researcher did not give any advice or kept count/time to enforce rhythm. For exercises where it suggested holding a position for 5 or 2 seconds, user was instructed at the beginning to keep count by themselves to preserve their natural rhythm. Our goal was to capture individual nuances of each user which replicates a scenario where a patient performs these exercises at home without the guidance of the physiotherapist. 

\section{Data visualisation}
\label{sec:vis}
In this section we visualise collected data, first visualise raw data from each sensor, secondly visualise the PM data using two dimensionality reduction methods. 

\subsection{Raw data}
\begin{figure}[ht]
  \centering
  \subfloat[Accelerometer data - Thigh and Wrist] {\includegraphics[width=1.0\textwidth]{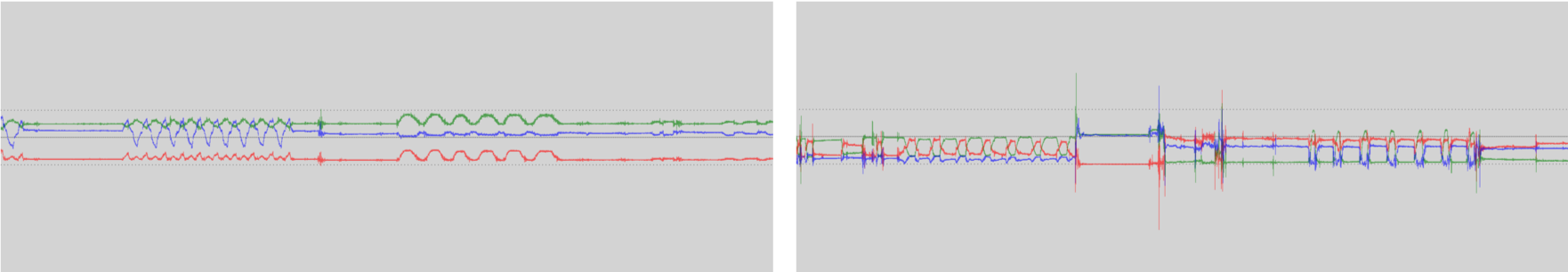}\label{fig:raw_ac}}
  \hfill
  \subfloat[Depth camera video]{\includegraphics[width=1.0\textwidth]{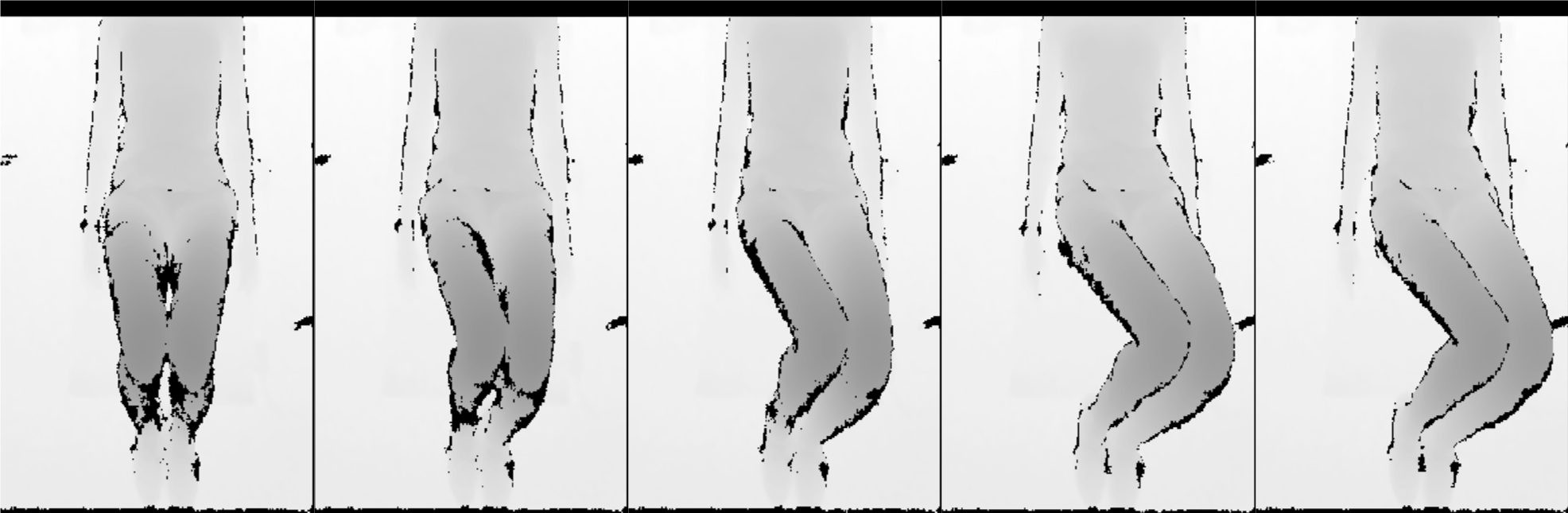}\label{fig:raw_dc}}
  \hfill
  \subfloat[Pressure mat data]{\includegraphics[width=1.0\textwidth]{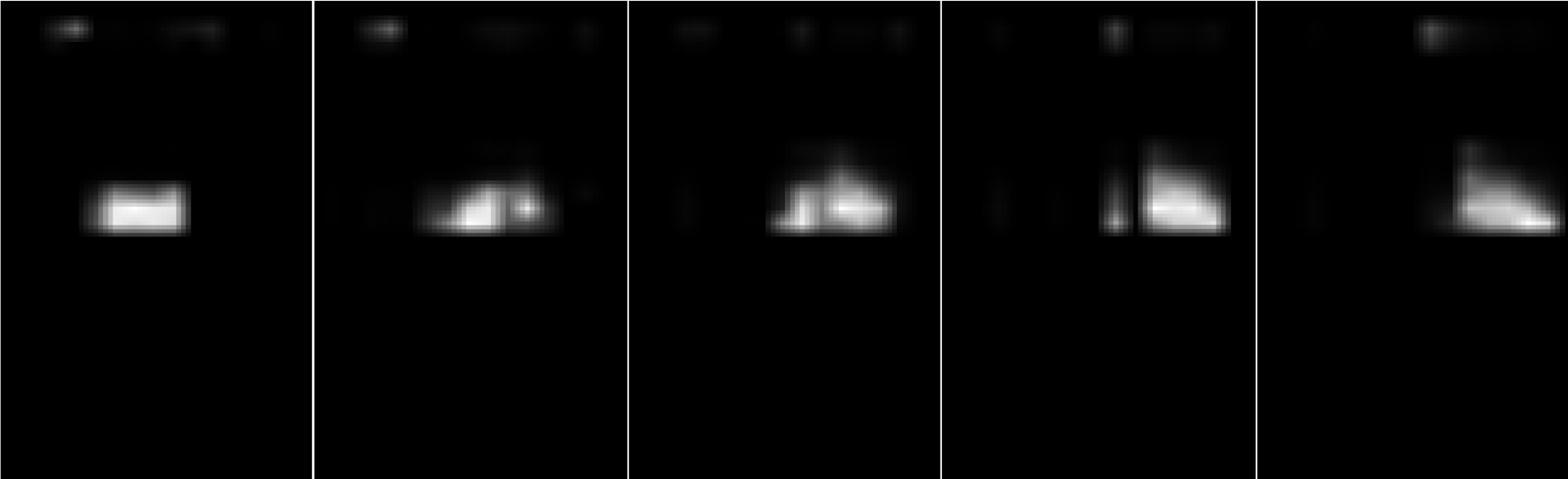}\label{fig:raw_pm}}
  \caption{Raw Data visualisation}
\end{figure}

Figures~\ref{fig:raw_ac} is the visualisation of thigh and wrist accelerometer sensor data over 5 exercises. It is evident that noisy outliers are found more commonly in the wrist sensor compared to the thigh sensor. It is also evident that some exercises does not require any movement or only move lightly of limbs which makes recognising exercises from a single accelerometer sensor challenging. Figure \ref{fig:raw_dc} show five depth camera data frames within 2 seconds from the Knee-rolling exercise and Figure~\ref{fig:raw_pm} show data frames captured by the pressure mat for the same exercise at the same timestamps. Depth camera visibly capture large amount of data compared to the pressure mat for Knee-rolling exercise, in-contrast an exercise such as Pelvic tilt seemed to be better captured with the pressure mat.

\subsection{PCA visualisation}
\begin{figure}[ht]
\centering
\includegraphics[width=1.0\textwidth]{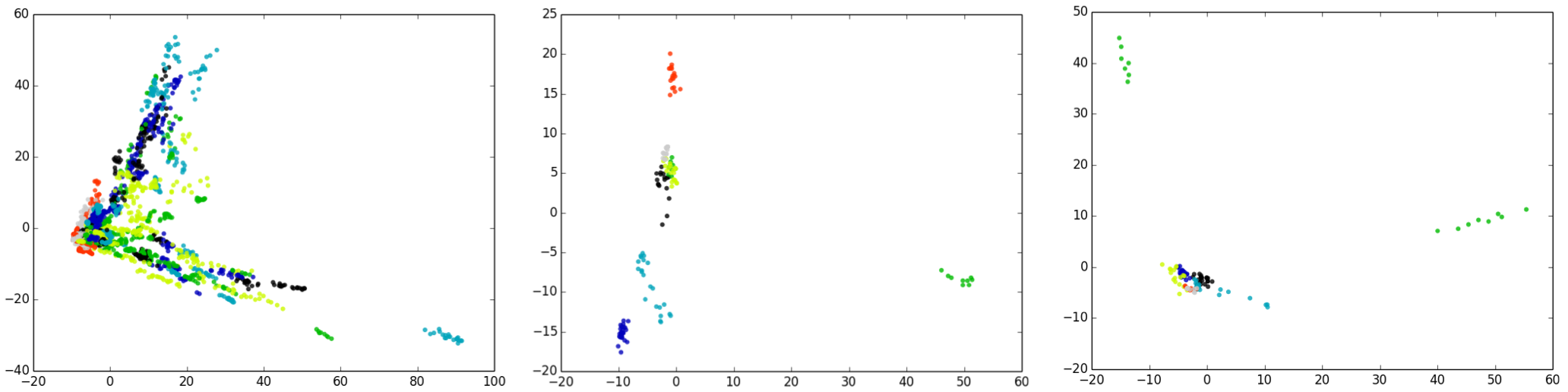}
\caption{PCA 2D embedding space of all user data, a physiotherapy user data and a regular user data} \label{fig:pca2}
\end{figure}

First we use Principal Component Analysis (PCA) for dimensionality reduction. We visualise the first 2 components clustered by exercise class in Figure~\ref{fig:pca2}. We observe that the cluster separation of data from a physiotherapy user is fairly significant compared to data from a regular user.

\subsection{t-SNE visualisation}
\begin{figure}[ht]
\centering
\includegraphics[width=1.0\textwidth]{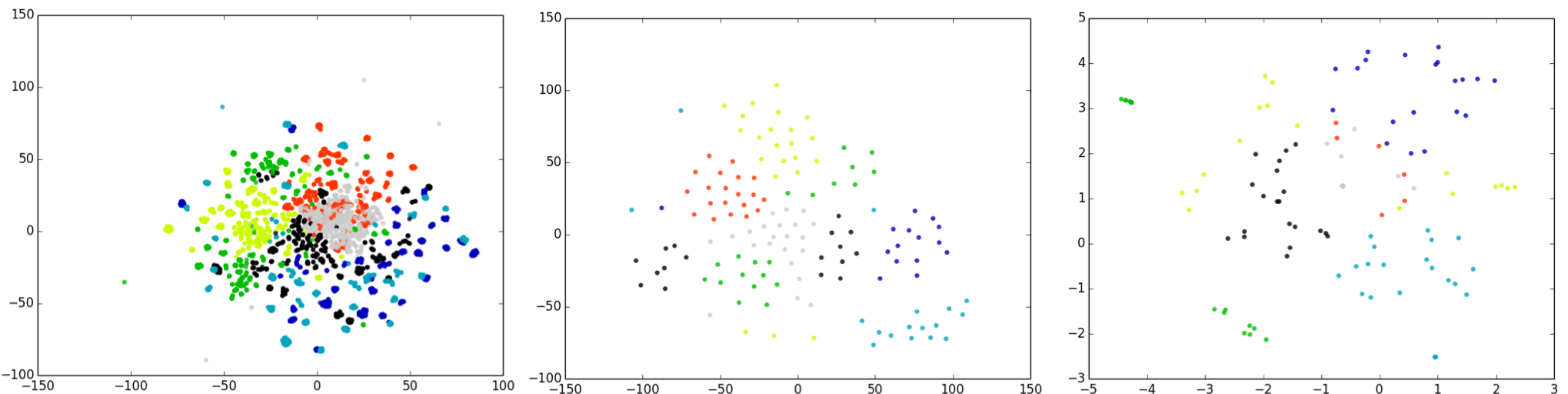}
\caption{t-SNE 2-D embedding space of all user data, a physiotherapist user data and a regular user data} \label{fig:tsne2}
\end{figure}

Secondly we select high dimensional data visualisation method t-Distributed Stochastic Neighbour Embedding(t-SNE). t-SNE build probability distributions over data pairs and maximise probability if pairs are similar and minimise probability if pairs are different. It uses Euclidean distance as the similarity measure. Figure~\ref{fig:tsne2} illustrates the t-SNE dimensionality reduction to two components clustered by exercise class, for all users, for a physiotherapy user and a regular user. Similar to PCA and more significantly, we observe that there are clear cluster boundaries in the physiotherapy user data and that it is difficult to capture clear cluster boundaries with a regular user data. 

Visualisation of PM data emphasise that automated learning of features is more suited for visual data compared to manual feature engineering or raw data. The visualisations also suggests that the potential of PM data to capture exercise performance quality. With this insight, PM data presents a great potential to explore qualitative assessment of exercise performance, that is essential to determine how a patient deviate from the correct execution once they are away from the supervision of the physiotherapist.

\section{Exercise Recognition with \mex}
\label{sec:eval}
In this section we present experiments that are designed to get reference performances on each individual sensor with standard machine learning (ML) and deep learning (DL) algorithms with pre-processing steps, algorithm details and experiment designs. 

\subsection{Pre-processing}
First few data cleaning steps were performed on the sensors that were continuously recording to remove the data that were recorded in between exercises. Then we have a dataset where data from four sensor are annotated with respective activity and user at each time-stamp.

To prepare this dataset for ML algorithms we use windowing method with overlap. We use 5 second window and 3 second overlap in these experiments and the values are kept constant across all sensors for comparability and consistency.  

We used a reduced frame rate of 1 frame/second with DC and PM data. The frame was selected by increasing the time-stamp by 1 second and selecting the nearest frame (i.e. in contrast to averaging the frame over the one second). Additionally DC data frames were compressed using OpenCV resize library from $240 \times 320$ to $12 \times 16$ considering computational memory requirements. Resulting raw data feature vector for DC or PM is of size $window \times frames\_per\_second \times frame\_width \times frame\_height$. An exploratory study of the hyper-parameters window size, overlap, frame rate, frame selection methods and frame sizes with regards to PM data can be found on the Appendix.

Similar windowing method was applied for accelerometer data ACW and ACT followed by DCT transformation. DCT transformation was applied to each axis individually and top 60 DCT components were selected. All DCT feature vectors were appended together resulting in a final vector of length 180. 
Table~\ref{tbl:meta} details meta-data of the pre-processed dataset. All are mean values averaged across number of folds or number of users.

\begin{table}[ht]
\caption{Meta-data}
\label{tbl:meta}
\centering
\ra{1.1}
\begin{tabular}{@{\hspace{5pt}} l @{\hspace{15pt}} l @{\hspace{15pt}} l @{\hspace{5pt}}}
\hline
\multicolumn{2}{c}{Property}&Value\\
\hline
\multicolumn{2}{l}{Total number of instances} & 6262 \\
\multicolumn{2}{l}{Instance per user} & 208 \\
\multicolumn{2}{l}{Instance per class per user} & 30 \\
\hline
\multirow{2}{*}{5-user fold cross validation}&Training instances & 5218 \\
&Test instance & 1043\\
\hline
\multirow{3}{*}{Input vector sizes} & ACT/ACW & $3 \times 60 = 180$ \\
 & DC & $5 \times 1 \times 12 \times 16 = 960$ \\
 & PM & $5 \times 1 \times 32 \times 16 = 2560$ \\
\hline
\multirow{2}{*}{DC/PM} & Minimum value & 0.0 \\
& Maximum value & 1.0 \\
\hline
\multirow{2}{*}{ACT/ACW raw data} & Minimum value & -8.0 \\
& Maximum value & +8.0 \\
\hline
\end{tabular}
\end{table}

\subsection{Evaluation Methodology}
Each experiment was evaluated with 5-user fold cross validation creating 6 folds. Each fold was repeated for 10 times for algorithms that are non-deterministic. User specific setting will use data from 25 of the 30 users in training 5 users in testing. This methodology emulates a real-life deployment setting where the end-user data are not seen during training.

Macro F-measure is the selected performance measure for the experiments as it provides a better representation of precision and recall compared to accuracy. F-measure is calculated for each label $i$, and their non-weighted mean is presented as the final value (Equation~\ref{eq:f1}). Weighted F-measure is not required in these experiment as the dataset is class balanced (i.e. contains similar amount of data instances for each class). 

\begin{gather}
F_1 = \sum_i 2 \times \frac{precision_i \times recall_i}{precision_i + recall_i}\label{eq:f1}
\end{gather}

\subsection{Experiments}
\label{sec:ex}
Here we list the classification algorithms and the different feature representations considered in the experiments. 

\begin{table}[ht]
\caption{Classification Algorithms}
\label{tbl:cls}
\centering
\ra{1.1}
\begin{tabular}{@{\hspace{5pt}} l @{\hspace{15pt}} l @{\hspace{5pt}}}
\hline
Algorithm&Description\\
\hline
kNN&K-Nearest Neighbours algorithm; we present results with k=1 and k=3\\
SVM&Support Vector Machine classifier with a Radial Basis Function kernel\\
\multirow{2}{*}{MLP}&Multi-layer Perceptron classifier with a softmax activation layer that \\&selects the maximum probability class as the predicted class. \\
\hline
\end{tabular}
\end{table}

\subsubsection{Feature representations}
\label{sec:fr}
\begin{figure}[t!]
\centering
\includegraphics[width=0.8\textwidth]{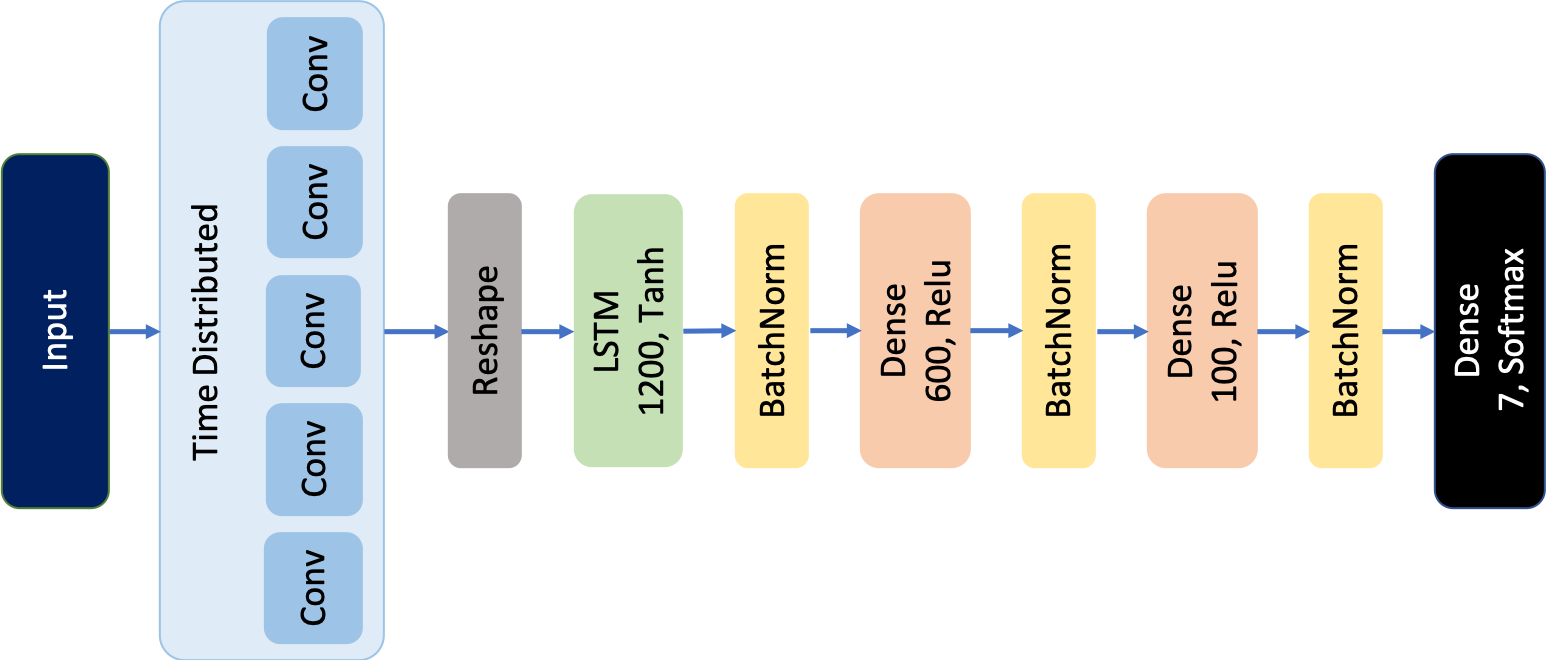}
\caption{LSTM model with time distributed convolutional feature embedding function} \label{fig:tdlstm}
\end{figure}

\begin{description}

\item [Raw:] Raw sensor data (flattened if necessary) as the feature representation. 

\item [DCT:] For ACT and ACW; the three axial time-series data are converted in a DCT feature vector by converting each axis to DCT feature vector, then selecting the first 60 elements of each vector and finally appending three vectors together to form a feature vector of length 180.  

\item [AE:] For PM and DC; An Auto-encoder (AE) with 5 hidden layers (Figure~\ref{fig:ae}) is trained with sensor data to reconstruct itself and the centre hidden layer with lowest dimension is used as the feature representation. Here the AE performs a dimensionality reduction and produce a feature vector of size 64 as the input of the classifier.
\begin{figure}[ht]
\centering
\includegraphics[width=0.6\textwidth]{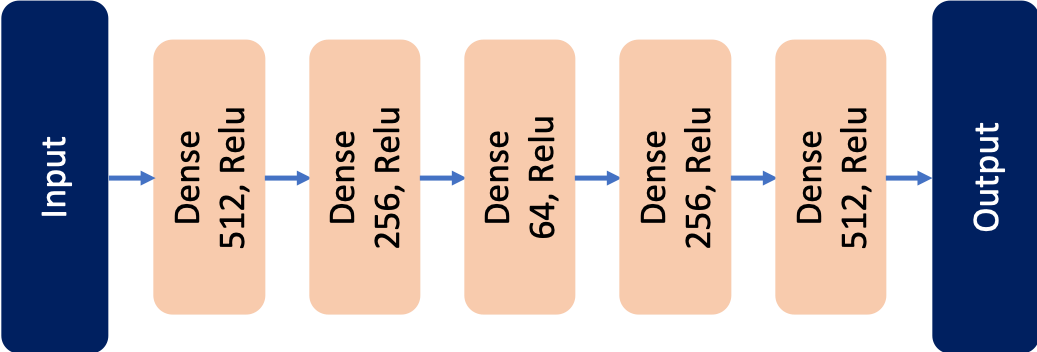}
\caption{Auto-encoder model} \label{fig:ae}
\end{figure}

\item [ANN:] Artificial Neural Network;  Artificial Neural Network, comprised of single or multiple layers of densely connected hidden layers. Each hidden layer is followed by a Batch Normalisation layer to normalise the output and to avoid over-fitting~\cite{ioffe2015batch}. Both variations below output a feature vector of size 100 as the input of the classifier.
\begin{itemize}
    \item Shallow-ANN: Consist of one hidden layer densely connected(Figure~\ref{fig:shallowann}). 
    \item Deep-ANN: Consist of five hidden layers as in Figure~\ref{fig:deepann}.
\end{itemize}

\begin{figure}[t!]
  \centering
  \subfloat[Shallow ANN model] {\includegraphics[width=0.3\textwidth]{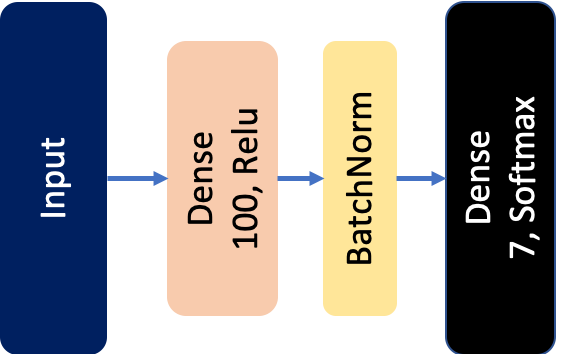}\label{fig:shallowann}}
  \hfill
  \subfloat[Deep ANN model]{\includegraphics[width=0.95\textwidth]{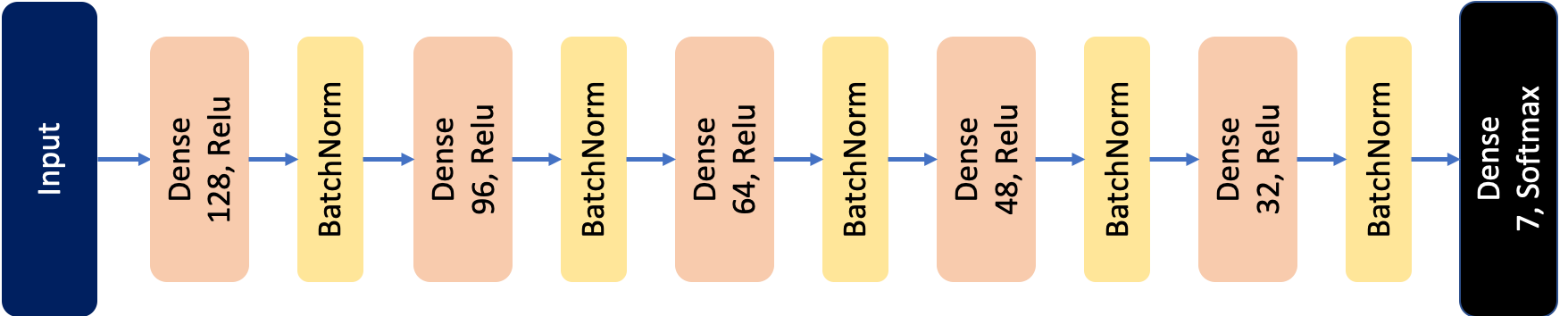}\label{fig:deepann}}
  \caption{ANN models}
\end{figure}

\item [CNN:] Convolutional Neural Networks. Similar to ANN, we use Batch Normalisation at the output of each hidden layer for regularisation. We explore three variations to suit different sensor modalities as follows and they all produce an output vector of size 100 as the input of the classifier. 
\begin{itemize}
    \item DCT-1D: For ACT and ACW; 1-dimensional convolutions and max pooling where the number of channels is 1 and the input feature length is 180. (Figure~\ref{fig:1dconv}).
    \item Raw-1D: Comprised of 1-dimensional convolutions and max pooling layers as in Figure~\ref{fig:1dconv}.
    For ACT and ACW, input is raw data of length 500 ($window \times frames\_per\_second$) with 3 channels (i. e. for 3 axes in accelerometer data). 
    For PM and DC data, each frame is flattened to form a vector and frames from a time window are appended together to crate the single dimension input feature vector of length $frame\_width \times frame\_height \times window \times frames\_per\_second$ with 1 channel.
    \item 2D: Consist of 2-dimensional convolution and max pooling layers(Figure~\ref{fig:2dconv}). For PM and DC data, with in a time window, frames are appended to form a 2D vector with 1 channel.
\end{itemize} 

\begin{figure}[t!]
  \centering
  \subfloat[1D convolution model] {\includegraphics[width=1.0\textwidth]{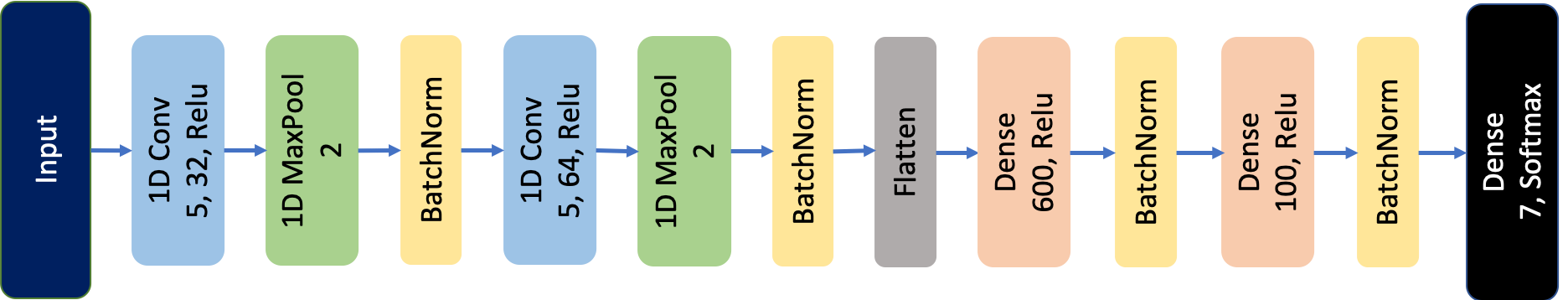}\label{fig:1dconv}}
  \hfill
  \subfloat[2D convolution model]{\includegraphics[width=1.0\textwidth]{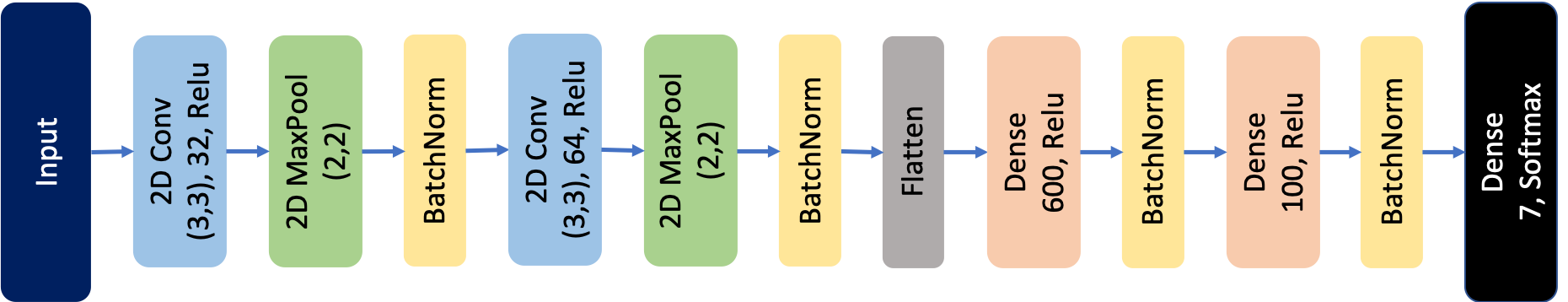}\label{fig:2dconv}}
  \caption{CNN models}
\end{figure}

\item [LSTM:] Long-short Term Memory Neural Networks. Similar to ANN and CNN, e use Batch Normalisation for regularisation. Here, time Distributed CNN architectures that learn low-level feature embeddings are followed by a LSTM layer that learns temporal dependencies as in Figure~\ref{fig:tdlstm}. Again we explore three variations as follows.
\begin{itemize}
    \item DCT-1D-CNN: Time distributed 1D Convolution architecture (similar to Figure~\ref{fig:1dconv}) for the DCT data of ACT and ACW. 
    \item Raw-1D-CNN: Time distributed 1D Convolution architecture (similar to Figure~\ref{fig:1dconv}) suited for all sensor modalities. 
    \item 2D-CNN: Time distributed 2D Convolution architecture (similar to Figure~\ref{fig:1dconv}) for PM and DC data. 
\end{itemize} 
\end{description}

All k-NN and SVM models were implemented using sklearn Python libraries~\footnote{https://scikit-learn.org/stable/} and the DCT transformation was done using scipy Python library~\footnote{https://docs.scipy.org/doc/scipy/reference/fftpack.html}. 
All parametric models were implemented with Keras with tensorflow backend, trained for 50 epochs, using Categorical Cross-entropy as the loss function. AdaDelta Optimiser\footnote{https://arxiv.org/abs/1212.5701} with default settings was used to optimise the model and training batch size was set to 32. All the pre-processing and model implementation code is publicly available on GitHub~\footnote{https://github.com/anjanaw/MEx}. 

In addition to above experiments an exploratory study of the dataset is included in Appendix.
First subsection in the Appendix compare each sensor in personalised and non-personalised evaluation settings. With these experiments we observe how each sensor performs with the introduction of end user data during training. Following subsections explore four hyper-parameters we considered during the pre-processing of the dataset; first we explore different window and overlap values with PM data followed by different frame rates and compression ratios with the PM data. These studies gained us insights in to how to improve performance that will be explored in future.

\section{Results}
\label{sec:results}
Table~\ref{tbl:results} summarises results we obtained with each sensor for the algorithms detailed in Section~\ref{sec:ex}.
For each sensor, the best performing algorithm is highlighted with bold text and algorithms that are significantly similar (t-test with 95\% confidence) to the best performing algorithm are highlighted with an asterisk. 
\begin{table}[t!]
\ra{1.2}
\centering
\begin{tabular} {@{\hspace{5pt}} l @{\hspace{5pt}} r @{\hspace{5pt}} r @{\hspace{15pt}} r @{\hspace{15pt}} r @{\hspace{15pt}}r @{\hspace{15pt}} r @{\hspace{5pt}}}
\hline
Classifier & \multicolumn{2}{c}{Embedding} & ACT & ACW & DC & PM\\
\hline
\multirow{3}{*}{1-NN}& \multicolumn{2}{c}{DCT}&0.7668&0.4868&- &-\\
& \multicolumn{2}{c}{Raw}&-&-&0.6937&0.5581\\
& \multicolumn{2}{c}{AE}&-&-&0.7426&0.6312\\
\hline
\multirow{3}{*}{3-NN}&\multicolumn{2}{c}{DCT}&0.7750&0.4928&-&-\\
& \multicolumn{2}{c}{Raw}&-&-&0.6824&0.5525\\
& \multicolumn{2}{c}{AE}&-&-&0.7454&0.6393\\
\hline
\multirow{3}{*}{SVM}&\multicolumn{2}{c}{DCT}&0.8578&0.4904&-&-\\
&\multicolumn{2}{c}{Raw}&-&-&0.7285&0.3760\\
&\multicolumn{2}{c}{AE}&-&-&0.7592&0.6937\\
\hline
\multirow{8}{*}{MLP}& \multirow{2}{*}{ANN}&Shallow&0.8726&0.5741&0.6377&0.6744\\
&&Deep&0.8521&0.5427&0.6796&0.6609\\
\cline{2-7}
&\multirow{3}{*}{CNN}&DCT-1D &0.8704&0.5754 &- &-\\
&&Raw-1D&0.7594&0.4245&0.8182&0.6902\\
&&2D &-&-&\textbf{0.8634}&0.7049*\\
\cline{2-7}
&\multirow{3}{*}{LSTM}&DCT-1DCNN &\textbf{0.8892}&\textbf{0.6499}&-&-\\
&&Raw-1DCNN&0.7026&0.4258&0.8285&\textbf{0.7194}\\
&&2DCNN&-&-&0.8056&0.7038*\\
\hline
\end{tabular}
\caption{Results}
\label{tbl:results}
\end{table}

In general MLP classifier achieved the best performance with different feature embedding functions. In addition, deep models performed comparatively better than the shallow models consistently. PM and DC sensors with visual data found 2D convolutional feature embedding functions are advantages compared to 1D feature embedding functions or learning temporal dependencies. In contrast ACW and ACT sensors with time-series data found learning temporal dependencies with 1D feature embedding to be more advantages.

ACT and ACW data are best represented for classification with the LSTM model with 1D-CNN as the feature embedding function. It is evident that learning temporal dependencies with LSTM models results in better feature representations for accelerometer data but it is also noteworthy that 1D-CNN models and ANN representations performs closely with the LSTM models. In addition, a notable observation is made between DCT and raw accelerometer data performance with deep models. Specifically, ACT and ACW DCT feature representations significantly outperform similar model with raw data 18.66\% and 22.41\% respectively for LSTM models and by 11.10\% and 15.09\% respectively for CNN models. These results confirm that DCT feature transformation achieves better performance compared to raw data for accelerometer sensors. This is on par with evidence seen in literature comparing raw vs. lazy feature engineering methods~\cite{sani2017learning}. 

Considering PM and DC raw data, kNN and SVM performs poorly suggesting the importance of learning a feature representation for visual data. Accordingly we compare the classification performance of raw data against a feature representation learnt with auto-encoders (AE). Feature representation learnt at the narrow hidden layer of the AE through dimensionality reduction improved performance significantly for both DC and PM data. With 3-NN algorithm, DC and PM data achieved 6.30\% and 8.68\% performance improvements respectively and with SVM it was 3.06\% and 31.77\%.

With DC data, best performance was observed with 2D-CNN model which significantly outperformed all other models. LSTM models with 1D-CNN feature embedding performs second best with DC data. PM data was best represented by the LSTM model with 1D-CNN feature embedding function. But it is noteworthy that 2D-CNN model and 2D-CNN-LSTM models performs significantly well.
These results suggest that the visual sensor data with are best presented for classification with convolution models. In general we observe that learning temporal dependencies with LSTM effectively contribute towards increased performance with PM and DC data. 

In summary these results emphasise the characteristics of different classification models and feature representation methods with different sensor data types. In future, we plan to optimise these models for single sensor classification as well as in multi-sensor fusion classification. 

\section{Conclusions}
\label{sec:conc}
This paper presents the MEx: Multi-modal Exercises Dataset for Human Activity Recognition and benchmark performance on standard classification algorithms.
The dataset contains 7 exercises recorded with four sensors, a depth camera, a pressure mat and two accelerometers. These sensors generate different data types, and we explore how classifier performance is affected by different shallow and deep feature representations of sensor data.
Our study on concluded that visual data such as pressure mat or depth camera data are better represented with 2-D convolutional architectures(2D-CNN) while, time-series data from accelerometers preferred the combination of shallow feature transformations(DCT) and learning temporal dependencies with recurrent architectures(LSTM).
With these results, we plan to explore multi-modal sensor fusion methods with attention mechanisms to find algorithms with improved performance towards implementing a exercise recognition system for patients with Musculoskeletal Disorders.
In addition the data visualisation suggests that the exercise performance quality is distinctly captured by pressure mat data which we will exploit in future to guide users to perform exercises with precision.

\section{Acknowledgement}
This work is part funded by SelfBACK. The SelfBACK project is funded by the European Union's H2020 research and innovation programme under grant agreement No. 689043. More details available at http://www.selfback.eu.

\bibliographystyle{splncs04}
\bibliography{ref}
\clearpage
\section*{Appendix}
\subsection*{Personalised vs. Non-personalised settings}
\label{sec:appA}
We explore two settings, personalised and non-personalised to measure the susceptibility of each sensor to personal nuances. 

\begin{itemize} 
\item Personalised: Classification models is tested on the same user set used during training.
\item Non-personalised: Model is trained and tested on data from mutually exclusive two sets of users.
\end{itemize}

Non-personalised evaluation is the established method of evaluation for HAR models as it resembles real-life scenario where end-users(test users) are not seen during training. With Personalised setting, we plan to observe the impact of introducing end user data to the learning process. 
For these experiments we used the LSTM model with 1D-CNN embedding for ACW and ACT data (achieved best performance in Section~\ref{sec:results}), and the 2DC-CNN model for DC and PM data.
We follow the same methodology here from Section~\ref{sec:eval} by repeating each 6-fold test 10 times. Mean f-measure was measured and presented in Table~\ref{tbl:np_p}. 

\begin{table}[ht]
\ra{1.4}
\centering
\begin{tabular} {@{\hspace{5pt}} l @{\hspace{15pt}} r @{\hspace{15pt}} r @{\hspace{15pt}} r @{\hspace{15pt}} r @{\hspace{5pt}}}
\hline
\multirow{2}{*}{Setting} & \multicolumn{4}{c}{Sensor}\\
\cline{2-5}
&ACT&ACW&DC&PM\\
\hline
Personalised & 0.9860&0.9277&0.9957&0.9743\\
Non-personalised &0.8879&0.6499&0.8634&0.7049\\
\hline
Improvement(\%)&9.80&27.78&13.22&26.94\\
\hline
\end{tabular}
\caption{Personalised vs. non-personalised settings} 
\label{tbl:np_p}
\end{table}

Results suggest that using end user data during training achieves significant performance improvement with all sensors.
Highest improvements are observed with ACW data of 27.78\% and PM data of 26.94\% followed by DC and ACT with 13.22\% and 9.80\% respectively. 

The results suggests that it is highly advantages for PM and ACW data models to learn from end-user data. It is also observed that these sensors performed poorly in non-personalised setting. It suggests that the feature embedding function learnt with training users is finding it is difficult to represent test-user data correctly (i.e. unable to generalise to new data), and once some of the end-user data is introduced in training, the models perform significantly better. These observations concludes that PM and ACW data carry the highest amount of personal nuances, resulting in highest differences between data from different users. 

In summary, a sensor such as PM is most inclined to capture personal nuances such as weight and shape; ACW sensor may capture more personal quirks and habits. In contrast data from sensors such as ACT or DC are indifferent to personal characteristics and are able to capture movements generalisable across users. This is an essential insight for when selecting sensors beyond the task of activity recognition, such as exercise quality assessment or counting where capturing personal characteristic is essential. 

\subsection*{Window and overlap}
\label{sec:appB}
An empirical study was conducted to understand the affect on classification performance with different window and overlap settings. We conducted a set of experiments with the PM data on the MLP classifier with 2D-CNN feature embedding function. 3, 5 and 8 seconds were considered as window sizes and a number of overlap values were considered for each window size. Experiments were done in a non-personalised setting and each 6-fold experiment was repeated 5 times and mean F1-measure is presented in Table \ref{tbl:w_o}. Best performing overlap for each window setting is highlighted in bold.

\begin{table}[ht]
\ra{1.3}
\centering
\begin{tabular} {@{\hspace{5pt}} l @{\hspace{15pt}} r @{\hspace{15pt}} r @{\hspace{15pt}} r @{\hspace{15pt}} r @{\hspace{15pt}} r @{\hspace{15pt}} r @{\hspace{5pt}}}
\hline
\multirow{2}{*}{Window Size} & \multicolumn{6}{c}{Overlap}\\
\cline{2-7}
&6&4&3&2&1&0\\
\hline
3&-&-&-&0.6500&0.6518&\textbf{0.6697}\\
5&-&-&0.7049&0.6992&\textbf{0.7167}&0.7175\\
8&0.7531&0.7514&-&\textbf{0.7666}&-&0.7433\\
\hline
\end{tabular}
\caption{Window and Overlap with PM data} 
\label{tbl:w_o}
\end{table}

Overall we observe that classification performance improves with the window size, which is intuitive given that with larger window size, the data instance carry more information to recognise an activity. 
But there are limitations to using larger window size in practice, firstly larger window size results in large data instances and smaller number of training instances. Limited number of training instances may cause parametric models to over-fit or not optimise. In addition, at test time, the time elapsed between two predictions is higher, which is not user friendly for real-time use. Therefore it is important to find a window and overlap size that are both practical in real-world applications while preserving classification performance.

\subsection*{Frame Selection, Frame Rate and Compression}
\label{sec:appC}
Achieving highest performance with small amount of data can be challenging due to information lost. An empirical study was carried out to observe the impact on performance when frames are down-scaled to reduce frame sizes and with reduced frame rates. Number of experiments were designed exploring three frame sizes and three frame rates with following frame selection techniques.

\begin{itemize} 
\item Average (AVG): A new frame is created by pixel wise averaging all frames within the time-period. 
\item Increment (INC): Increment time-stamp by the time-period and select the nearest time-stamp and respective frame.
\end{itemize}

Each 6-fold experiment was repeated 5 times and mean F1-measure is presented in Table \ref{tbl:s_f}. Best performing overlap for each window setting is highlighted in bold.

\begin{table}[ht]
\ra{1.3}
\centering
\begin{tabular} {@{\hspace{5pt}} l @{\hspace{15pt}} c @{\hspace{15pt}} r @{\hspace{15pt}} r @{\hspace{15pt}} r @{\hspace{5pt}}}
\hline
\multirow{2}{*}{Frame Size} & Frame Selection & \multicolumn{3}{c}{Frames per second}\\
\cline{3-5}
& Method & 1 & 2 & 3 \\
\hline
\multirow{2}{*}{1($32\times16$)} & INC & \textbf{0.7049}&0.6823&0.6979 \\
& AVG & 0.7012&0.6815&0.6941\\
\hline
\multirow{2}{*}{2($16\times16$)} & INC & \textbf{0.7021}&0.6697&0.6906 \\
& AVG & 0.6898&0.6882&0.6845 \\
\hline
\multirow{2}{*}{3($8\times8$)} & INC & 0.6685&0.6816&\textbf{0.6946}\\
& AVG & 0.6799&0.6902&0.6795 \\
\hline
\end{tabular}
\caption{Frame Selection, Rate and Compression Results with PM data}
\label{tbl:s_f}
\end{table}

Considering the frame size, we observe that both original frame size 1 and down-scaled frame size 2 perform similarly and the performance is degraded with frame size 3. 
It is also observed that increased frame rate does not necessarily contribute towards improved performance as frame sizes 1 and 2 achieve best performance with frame rate 1 frame per second.
But once frame size is reduced to $8\time8$ it is evident that increased frame rate improves accuracy. 
In general INC method yielded the better performance across all three frame sizes. But we observe no statistically significant performance difference between the AVG and INC methods. 

The important take away from this study is that frame size and frame rate can be selected such that they compensate for each other, and more importantly selecting a higher frame rate and a larger frame size does not naturally improve performance. Lazy feature augmentation techniques such as frame down-scaling and frame selection can be pivotal to achieving better performance with reduced memory and computational power. 
\end{document}